\documentclass{article}

     \usepackage[preprint,nonatbib]{neurips_2020}

\usepackage[utf8]{inputenc} % allow utf-8 input
\usepackage[T1]{fontenc}    % use 8-bit T1 fonts
\usepackage{hyperref}       % hyperlinks
\usepackage{url}            % simple URL typesetting
\usepackage{booktabs}       % professional-quality tables
\usepackage{amsfonts}       % blackboard math symbols
\usepackage{nicefrac}       % compact symbols for 1/2, etc.
\usepackage{amsmath}
\usepackage{graphicx}
\usepackage{multirow}
\usepackage{amssymb}
\usepackage{pifont}
\usepackage[nolist]{acronym}
\usepackage{subfig}
\usepackage{microtype}   

\usepackage[linesnumbered,ruled]{algorithm2e}

\title{A Novel Sampling Scheme for Text- and Image-Conditional Image Synthesis in Quantized Latent Spaces}

\author{Dominic Rampas\textsuperscript{\textsection}\\
Technische Hochschule Ingolstadt\\
Ingolstadt, Germany \\
and\\
Wand Technologies Inc.\\
New York, USA \\
{\tt\small dominic.rampas@gmail.com}
\And
Pablo Pernias\textsuperscript{\textsection}\\
Independent Researcher \\
Sant Joan d'Alacant, Spain \\
{\tt\small pablo@pernias.com}
\And 
Marc Aubreville \\
Technische Hochschule Ingolstadt\\
Ingolstadt, Germany \\
{\tt\small marc.aubreville@thi.de}
}

\newcommand{\etal}{\textit{et al}.~}

\begin{document}

\maketitle

\begin{abstract}
    Recent advancements in the domain of text-to-image synthesis have culminated in a multitude of enhancements pertaining to quality, fidelity, and diversity. Contemporary techniques enable the generation of highly intricate visuals which rapidly approach near-photorealistic quality. Nevertheless, as progress is achieved, the complexity of these methodologies increases, consequently intensifying the comprehension barrier between individuals within the field and those external to it.
    In an endeavor to mitigate this disparity, we propose a streamlined approach for text-to-image generation, which encompasses both the training paradigm and the sampling process. Despite its remarkable simplicity, our method yields aesthetically pleasing images with few sampling iterations, allows for intriguing ways for conditioning the model, and imparts advantages absent in state-of-the-art techniques. To demonstrate the efficacy of this approach in achieving outcomes comparable to existing works, we have trained a one-billion parameter text-conditional model, which we refer to as "Paella". In the interest of fostering future exploration in this field, we have made our source code and models publicly accessible for the research community.

\end{abstract}

\begin{figure}[ht]
\includegraphics[width=\textwidth]{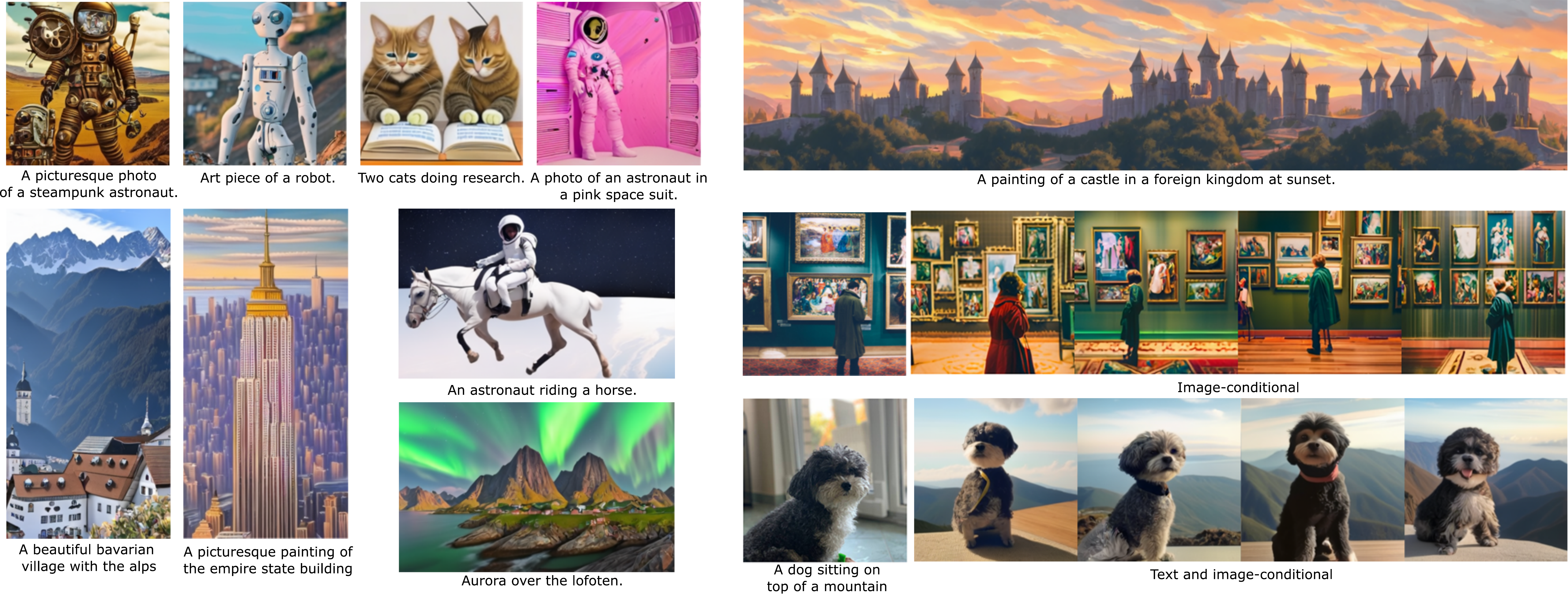}
\caption{Visual results of our proposed method and trained model. It is able to perform a variety of image synthesis tasks. The left hand side and the top right shows our model's abilities on text-conditional image generation on different sizes while the two bottom right panels show image- and combined text and image-conditioning. }
\label{fig:samples}
\end{figure}

%%%%%%%%% BODY TEXT
\section{Introduction}
\label{sec:intro}
Recent advancements in the field of text-to-image synthesis \cite{gafni2022make,dalle2_ramesh2022hierarchical,saharia2022photorealistic,rombach2022high,chang2023muse} have demonstrated substantial progress, as evidenced by the augmented diversity, enhanced quality, and the increased variation of generated images. At present, the majority of cutting-edge methodologies in this area primarily utilize either diffusion-based models \cite{ho2020denoising,saharia2022photorealistic,dalle2_ramesh2022hierarchical} or adopt transformer\cite{transformers_vaswani2017attention}-based architectures \cite{chang2023muse}. Although diffusion models achieve remarkable fidelity these outputs come with a heightened computational demand due to numerous sampling iterations and the amplifying complexity resulting from the application of increasingly sophisticated techniques. This intensification can lead to decreased inference speeds, rendering real-time implementation in end-user applications impractical. Further, it contributes to an elevated level of abstraction in comprehending the state-of-the-art technology, particularly for individuals outside of this specialized discipline. 
Although much work has been going into decreasing the number of sampling steps \cite{https://doi.org/10.48550/arxiv.2206.00364,ho2020denoising,https://doi.org/10.48550/arxiv.2202.09778}, this stays an open problem and many large text-to-images models relying on diffusion are still using rather high numbers of steps.
On the contrary, recent transformer-based approaches are able to use a much smaller number of inference steps, however employ a significant level of spatial compression, which can result in more artifacts but is necessary due to the self-attention mechanism growing quadratically with latent space dimensions. Furthermore, a transformer treats images as one-dimensional sequences by flattening the encoded image tokens, which is an unnatural projection of images and requires a significantly higher model complexity to learn an understanding of the 2D structure of images.  

\begin{figure}[t]
  \centering
  \includegraphics[width=\linewidth]{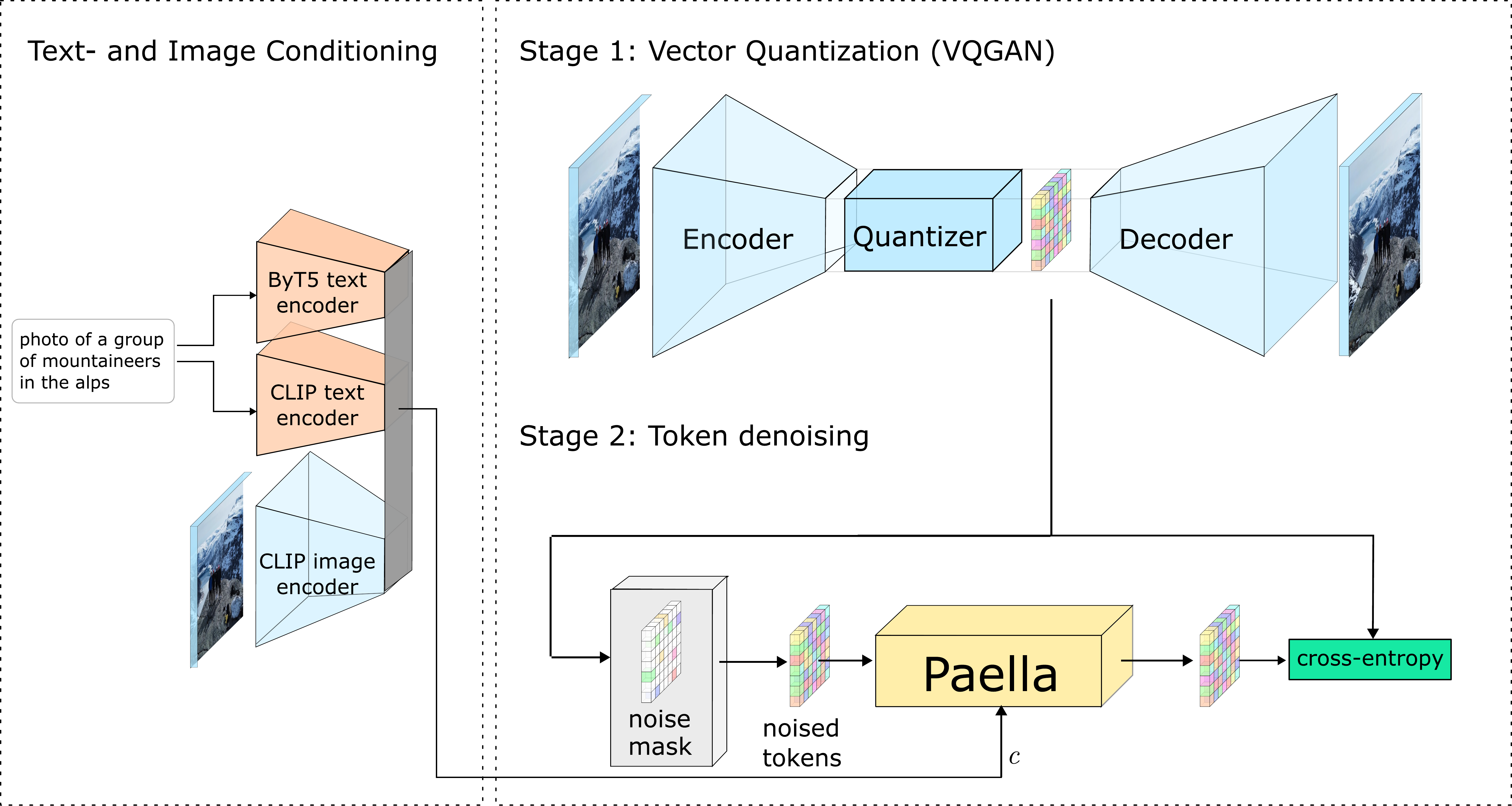}

  \caption{Visual depiction of the overall architecture of our proposed method. Training of \textit{Paella} operates on a compressed latent space. Latent images are noised and the model is optimized to predict the unnoised version of the image.}
   \label{fig:train_sample}
\end{figure}

\begin{figure}[ht]
\centering
\includegraphics[width=1.0\textwidth]{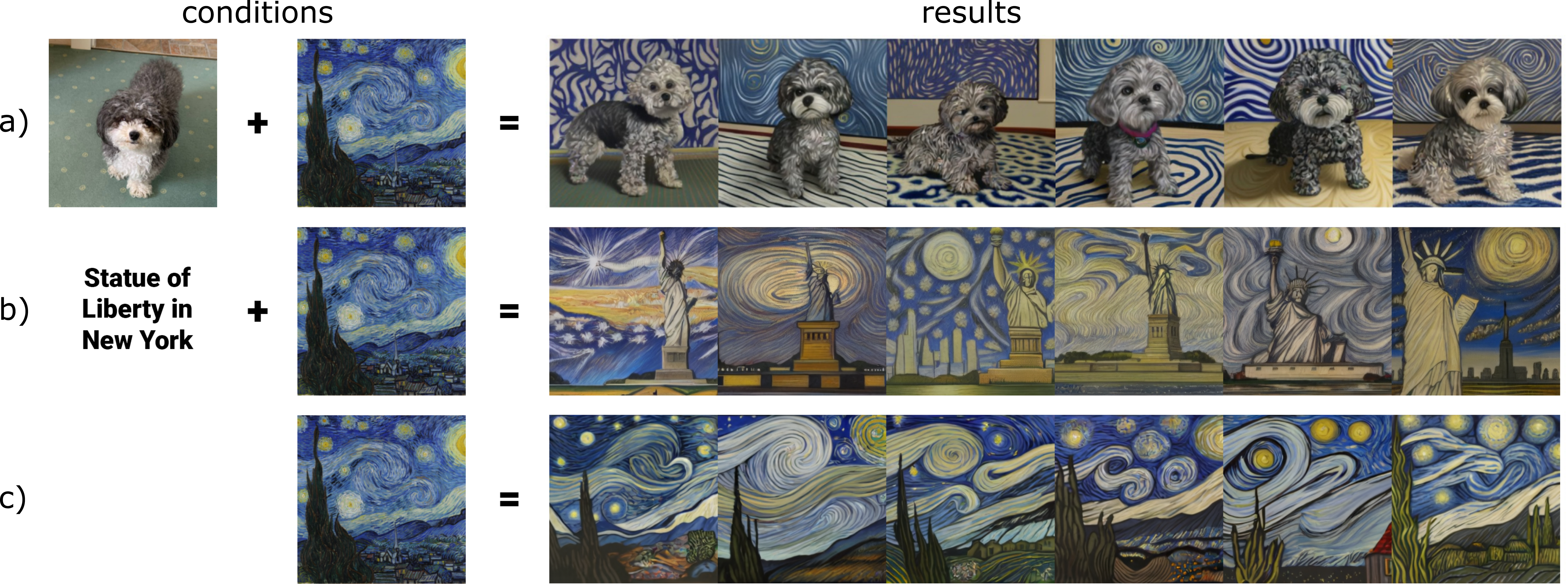}
\caption{Image manipulation with Paella. a) shows image-conditional image generation, b) shows text-conditional image generation and c) shows image variations.}
\label{fig:variations}
\end{figure}

In this work, we propose a novel technique for text-conditional image generation, diverging from both transformer and diffusion-based approaches. Our model enables image sampling with remarkable efficiency, requiring as few as 12 steps while consistently producing high-fidelity results. Moreover, the simplicity of our proposed technique enhances its accessibility, empowering individuals from various backgrounds to comprehend and implement this influential technology of text-to-image.
Our proposed model uses a convolutional paradigm and functions within a discretized latent space, utilizing a Vector Quantized Generative Adversarial Network (VQGAN) \cite{vqgan_esser2021taming} for both the encoding and decoding procedures (see Figure~\ref{fig:train_sample}) at a modest compression rate. Given the convolutional characteristics of our model, we are capable of operating at significantly reduced compression rates, thereby circumventing conventional transformer constraints like quadratic memory expansion.
Operating at a low compression rate facilitates the preservation of intricate details that are frequently altered in the context of higher compression scenarios. Throughout the training phase the quantized image tokens are noised by random token replacement. The model is subsequently tasked to reconstruct the image tokens, taking into account the noised variant and conditional input. The process of sampling new images unfolds iteratively and draws inspiration from methodologies employed by MaskGIT \cite{maskgit_chang2022maskgit} and MUSE \cite{chang2023muse}, but with significant changes:
Both previous approaches incorporate a unique mask token and employ it to initially obscure the entire image. Subsequently, the model makes iterative predictions for all tokens in the image concurrently, retaining only a select number of tokens about which the model exhibits the highest confidence, while the remaining tokens are masked once more.
We hypothesize this process to be inherently restrictive, as it precludes the model's capacity for self-correction of its early-stage predictions during sampling. To offer greater adaptability to the model, we randomly noise tokens as an alternative to masking them. This adjustment provides the model with the capability to refine its predictions for specific tokens throughout the sampling sequence.
We enable text conditioning by using ByT5-XL \cite{xue2022byt5} embeddings while additionally injecting pooled \ac{CLIP} \cite{clip_radford2021learning} text-, and image embeddings intermittently.
Owing to the convolution-based and spatially invariant nature of our token predictor, it possesses the capacity to generate images of any size, theoretically. In contrast, transformer-based models are required to incrementally adjust the context window to produce larger latent resolutions.
Furthermore, as the model is conditioned in part on image embeddings, it affords the ability to generate variations of images in a zero-shot context and to combine images with supplementary text prompts. (Figure \ref{fig:variations}).

Our main contributions are the following:
\begin{enumerate}
\item We present an innovative convolution-based model for text- and image-conditional image generation, distinctly diverging from prevalent transformer and diffusion-based methods, thus providing a fresh viewpoint in the image generation field.
\item Our model demonstrates impressive efficiency, requiring only 12 steps for high-fidelity image generation, and is designed to enhance accessibility, democratizing the comprehension and implementation of this technology across diverse academic and professional spectra.
\item We introduce a novel approach for text-to-image sampling in quantized latent spaces, employing token renoising, which is deviating from traditional methods that rely on a special mask token.  This enables iterative refinement of model predictions during the sampling sequence, thereby enhancing prediction accuracy.
\item We are publicly releasing the source code and the entire suite of model weights, all of which are licensed under the provisions of the MIT license.
\end{enumerate}

%-------------------------------------------------------------------------
\section{Related Work}
\label{sec:rel_work}
\subsection{Conditional Image Generation}
The field of text-conditional image generation has witnessed substantial advancements in recent periods. Initial explorations predominantly relied on \acp{GAN} \cite{reed2016generative,zhang2017stackgan}. More recent methodologies have seen the emergence of a unique image generation framework known as diffusion \cite{https://doi.org/10.48550/arxiv.1503.03585,ho2020denoising}, which has not only achieved parity with \acp{GAN} but has, in certain instances, surpassed them in both conditional and unconditional image generation \cite{dhariwal2021diffusion}. Diffusion models propose a score-based architecture that iteratively purges noise from a target image, with the training objective articulated as a reweighted variational lower-bound. Recent investigations have demonstrated that diffusion models can be effectively scaled to high resolutions through multi-stage strategies, whilst preserving the capacity to generate high-fidelity images \cite{rombach2022high,saharia2022photorealistic,dalle2_ramesh2022hierarchical}.

Recent advancements in image generation models \cite{rombach2022high,chang2023muse,vqgan_esser2021taming,gafni2022make,ramesh2021zero,ding2021cogview} frequently employ a two-stage approach, wherein the initial stage involves encoding images into a more condensed latent space. The second stage encompasses learning within this compressed latent space, an approach that has proven to be more efficient for training a text-conditional model due to its lesser computational demands compared to pixel-level training.
Initially, transformer-based models operated on an autoregressive basis, leading to a substantial deceleration in inference due to the necessity to individually sample each token. However, contemporary methodologies \cite{ding2022cogview2,maskgit_chang2022maskgit,chang2023muse} have adopted the use of a bidirectional transformer to mitigate the limitations inherent in autoregressive models. 
Consequently, image generation can be achieved using a reduced number of steps, while simultaneously leveraging a global context during the generation process.

\subsection{Conditional Guidance}
Text conditional guidance of models is usually achieved by encoding text prompts with a pretrained language model. Two principal categories of text encoders are widely utilized: contrastive text encoders and uni-modal text encoders.
\ac{CLIP} \cite{clip_radford2021learning}, a contrastive multimodal model, endeavors to align semantically analogous textual descriptions and images within a unified latent space. Numerous recent approaches for image generation have relied on a frozen \ac{CLIP} model as their sole method of conditioning. Dalle-2 by Ramesh \etal \cite{dalle2_ramesh2022hierarchical} only uses \ac{CLIP} image embeddings as input to their diffusion model, while relying on a ``prior'' converting \ac{CLIP} text embeddings to image embeddings. Stable Diffusion \cite{rombach2022high} employs un-pooled \ac{CLIP} text embeddings to condition its latent diffusion model \cite{rombach2022high}. 
Contrastingly, works by Saharia~\etal~\cite{saharia2022photorealistic}, Liu~\etal~\cite{liu2022character} and Chang~\etal \cite{chang2023muse} use a uni-modal large language model (T5~\cite{raffel2020exploring} or ByT5~\cite{xue2022byt5}) that can accurately encode textual prompts, resulting in more precise depictions regarding composition, style and layout.

\section{Method}
\label{sec:method}
\subsection{Training}
\begin{figure}[t]
  \centering
%   \fbox{\rule{0pt}{2in} \rule{0.9\linewidth}{0pt}}
  \includegraphics[width=0.8\linewidth]{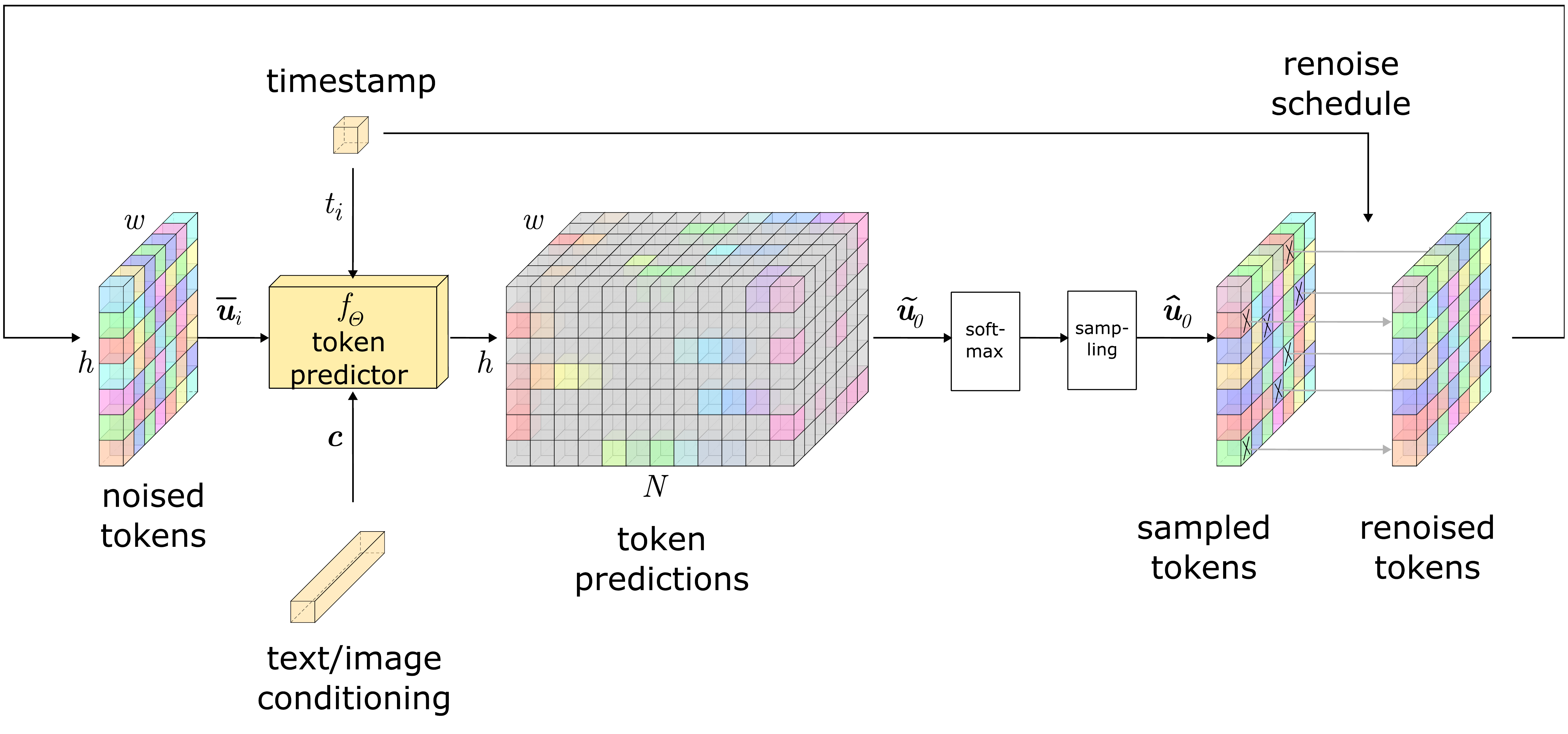}

  \caption{Sampling mechanism for the token predictor of our model.}
   \label{fig:sampling}
\end{figure}

Our proposal builds on the two-stage paradigm introduced by Esser \etal \cite{vqgan_esser2021taming} and consists of a \ac{VQGAN} for projecting the high-dimensional images into a lower-dimensional latent space, as shown in Figure~\ref{fig:train_sample}. Specifically, an encoder takes in the image at its base resolution of $H\times W\times C$ and maps it to a latent representation $\mathbf{u}$ with a resolution of 
$h\times w\times z$ with $h = H/f, w = W/f$, where $f$ is the compression rate. This operation is followed by a quantization step, discretizing the latents by
replacing each vector by its nearest neighbour from a learned codebook $Q \in \mathbb{R}^{N_{CB}\times z}$ of size $N_{CB}$. 
Afterwards, the quantized representation is given to the decoder which tries to reconstruct the input image. We use a pretrained \ac{VQGAN} with an $f=4$ compression and a base resolution of $256\times 256\times 3$, mapping the image to a latent resolution of $64\times 64$ indices.
The secondary phase involves learning the distribution of tokens within images situated in the low-dimensional latent space. During the training process, we introduce noise into the latent tokens derived from the encoded and quantized images by randomly substituting a certain proportion of the tokens with other randomly selected tokens from the codebook, as illustrated in Figure~\ref{fig:sampling}. The precise quantity of noised tokens is ascertained by randomly sampling from a uniform distribution ranging between 0 and 1, which the noising function interprets as the ratio of tokens to replace.
Notice, we don't use a specific scheduling function to determine the ratio of tokens to noise as done in \cite{maskgit_chang2022maskgit,chang2023muse}.
At training, a random number $t \sim \mathcal{U}(0, 1)$ is generated for each image. Subsequently, we sample a binary noise mask $\mathbf{m}$, with elements $m_{x,y}$, where a value of $m_{x,y} = 0$ indicates to keep the current token and a value of $m_{x,y} = 1$ signifies to noise it. The ratio of noised tokens is set to be equivalent to $t$. 
The noised tokens ${\bar{u}_{x,y}}$ at coordinate $(x,y)$ are subsequently drawn from a uniform distribution $n_{x,y} \sim \mathcal{U}(0,N-1)$ spanning all codebook indices:

$${\bar{u}_{x,y}} = \left\{ \begin{array}{rcl}u_{x,y}&\textrm{if}&m_{x,y} = 0\\n_{x,y} & \mathrm{else} \end{array} \right. $$

The noised image representation, denoted as ${\bar{u}}$, is subsequently inputted into the token prediction model $f_\theta$ in conjunction with two additional parameters: the timestep $t$ and a condition $c$. Mirroring the approach in diffusion models \cite{ho2020denoising}, we chose to incorporate the current timestep (noising ratio) into the model's input to provide an explicit information source concerning the amount of noise present within the image and the corresponding noise reduction expectation. While any condition can be included, such as class labels or semantic segmentation maps, we decided to condition the model on ByT5-XL \cite{xue2022byt5} as the main source 95\% of the time. Additionally, we also condition on \ac{CLIP} \cite{clip_radford2021learning} text-, and image-embeddings 5\% of the time. The noised indices, the timestep embedding and the conditioning are given as input to the token predictor model which is tasked to predict the un-noised tokens, yielding a prediction $\mathbf{\tilde{u}} \in \mathbb{R}^{h \times w \times N_{CB}}$.

\[\mathbf{\tilde{u}} = f_{\theta}(\mathbf{\bar{u}}, \mathbf{c}, \mathbf{t})\]

The token predictor is optimized via cross-entropy using label smoothing. Preliminary experiments indicated that for small timesteps the model tended to emulate the identity function. We hypothesize that this was due to the already small loss, as a result of not many tokens being subjected to noising. This led to noisy samplings as the final iterations of noise removal were not adequately learned. To circumvent this problem, we introduced a loss weighting schedule that reduces the contribution to the loss of tokens that were not noised for smaller timesteps. The loss weighting is defined as follows: 
\[l_w = 1 - (1 - m_{x,y}) \cdot ((1 - t) \cdot (1 - \eta))\]
We define $\eta$ to be the minimum value a token can take for the loss contribution. We found $m_v = 0.3$ to yield satisfactory results.

\begin{algorithm}
 \SetKwInOut{Input}{Input}
 \SetKwInOut{Output}{Output} 
 \Input{model $\mathbf{f_\theta}$, conditioning $\mathbf{c}$, latent shape $s$, sequence of noise ratios (timesteps) $t_1 > t_2 \cdot\cdot\cdot > t_{T}$, temperature $\tau$, cfg-weight $w$, codebook size $N_{CB}$}
\Output{tokens to be decoded using the VQGAN decoder $\hat{\mathbf{u}}$}
$\mathbf{\hat{u}_\mathrm{init}} \gets \mathrm{randint}(\mathrm{low}=0, \mathrm{high}=N_{CB}, \mathrm{shape}=s)$ \tcc*{Random initialization}
 
$\mathbf{\hat{u}} \gets \mathbf{\hat{u}}_\mathrm{init}$\;
 
 \For{$i=1$ \KwTo $T$}{
  $\tilde{\mathbf{u}} \gets \mathbf{f_\theta}(\hat{\mathbf{u}}, \mathbf{c}, t_i)$ \tcc*{Single inference step}
  
  $\tilde{\mathbf{u}} \gets \tilde{\mathbf{u}} \cdot w + \mathbf{f_\theta}(\hat{\mathbf{u}}, \mathbf{c}_\emptyset, t_i) \cdot (1-w)$ \tcc*{CFG-weighting}

   $\tilde{\mathbf{u}} \gets \mathrm{softmax}(\frac{\tilde{\mathbf{u}}}{\tau})$\;

   $\mathbf{\hat{u}} \gets \mathrm{multinomial}(\tilde{\mathbf{u}})$ \tcc*{Token multinomial sampling}

   \If{i<T}{
   $\hat{\mathbf{u}} \gets \mathrm{renoise}(\hat{\mathbf{u}}, t_{i+1}, \mathbf{u}_\mathrm{init})$ \tcc*{Token renoising}
   }
  
 }
 %return $\mathbf{\hat{u}}$
 \caption{Sampling function}
\label{algo:sampling}
\end{algorithm}

\subsection{Sampling}
While sampling in a single step would technically be possible, this procedure does not align properly with the training objective, as pointed out by \cite{maskgit_chang2022maskgit}. Therefore, we are using an iterative approach for sampling too. Let $\mathbf{u}_{T} \in \mathbb{N}_0^{h \times w} $ be a latent image where each value is a random token from the codebook. Furthermore, let $\mathbf{t} = \left[t_1, t_{2}, \dots, t_{T} \right]$ be the sequence of noising ratios at defined timesteps starting at $t_1=1$ (fully noised) and $t_{T}=0$ (noise-free) with $T$ being the number of sampling steps.  Moreover, $\mathbf{c} \in \mathbb{R}^{d}$ denotes the conditional embedding. Sampling is conducted in an iterative fashion and the following steps are executed in each iteration:
\begin{enumerate}
    \item The current noising ratio (time-step) $\mathbf{t}_i$, the latent space representation of the input $\mathbf{u}_{i}$, and the embedding $\mathbf{c}$ is given as input to the denoising model and it will predict all tokens simultaneously resulting in a score for each codebook index for the entire latent image. Specifically, after feeding an input $\mathbf{u}_{i}$, the output $\mathbf{\tilde{u}}_{0}$ has a shape of $h \times w \times N_{CB}$ where $N_{CB}$ is the number of codebook vectors.
    \item Afterwards a softmax function is applied to convert all scores to a probability distribution for each token in the latent image. Next, multinomial sampling is employed to sample one token from each distribution according to the probability. The result $\mathbf{\hat{u}}_{0}$ has a shape of $h \times w$.
    \item We randomly renoise a certain proportion of all sampled tokens back to their initial noise codebook values $\mathbf{\hat{u}}_\mathrm{init}$. This proportion is determined by the current noising ratio $\mathbf{t}_i$. 
\end{enumerate}
A visual depiction of the sampling algorithm can be seen in Figure \ref{fig:sampling} and the complete algorithm is given in Algorithm~\ref{algo:sampling}. Note that we renoise using the initial noise tokens instead of generating new random noise. We found this to lead to more robust outputs. Furthermore, unlike MaskGIT/MUSE we do not renoise the tokens with the lowest confidence and keep the ones with the highest scores, as this was not found to improve performance of our model. Instead, we renoised a random set of tokens for the sake of simplicity.

Additionally, we employed classifier-free-guidance (\ac{CFG}) \cite{classifier_free_guidance_ho2022classifier} for improving the sampling process. A null-label is introduced in the training. During sampling, we sample once with the null-label and once with the conditioning embedding. Afterwards we linearly interpolate between the logits (see line 5 in Algorithm \ref{algo:sampling}).
%$$\mathbf{u_{t}} = w \cdot \mathbf{u_{t, c}} + (1-w) \cdot \mathbf{u_{t, \emptyset}}$$
%where $w$ is a classifier weight that determines the pull towards the conditional sample.

\subsection{Token Predictor Design Choices}
\begin{figure}
\centering
\includegraphics[width=1.0\textwidth]{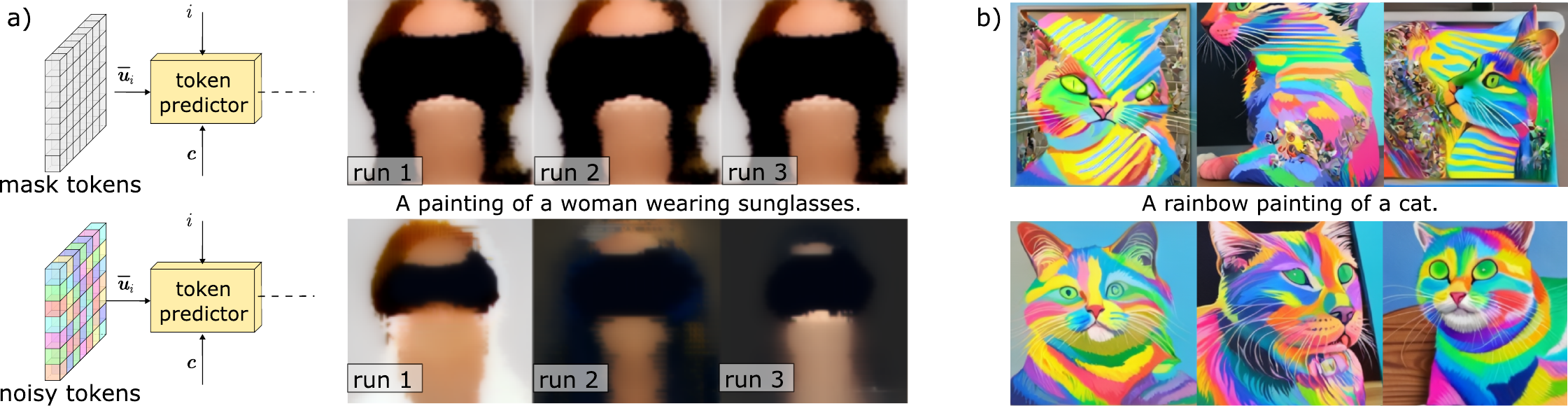}
\caption{a) Illustrative comparison between a single-step argmax denoising using masked tokens and random noise. The former (illustrated in the top row) always results in the same output, whereas random tokens give different outputs (bottom row), showing an intrinsic induced diversity in our method, while in a masked setting diversity needs to be induced in sampling. b) Comparison between low confidence renoising as used in MUSE (top) vs. our proposed random renoising (bottom).}
\label{fig:masking}
\end{figure}

% - talk about speed optimized arch
% - no attention used
Chang \etal use a bidirectional transformer for the image synthesis task. We argue that this implies two limiting factors: 1. Using a transformer necessitates treating the image as a flat 1D sequence, which is an unnatural projection of images and may impose a fundamental disadvantage during learning, since the 2D structure first needs to be learned through positional embeddings. 2. The quadratic memory growth limits transformers to small latent space resolutions, which in turn requires high compression rates. By replacing the transformer by a convolutional model, both aforementioned problems are resolved by having much lower memory requirements and induced 2D biases. 
Furthermore, we hypothesize the sampling strategy of MUSE/MaskGIT being too restrictive, as a token can not be changed after it has been fixed except by explicitly re-masking it, which prohibits the model to refine it's prediction at subsequent sampling iterations.
Unfortunately, MUSE has no public release of the code \& weights, making it hard to prove this hypothesis. To compensate this, we employed their sampling strategy and use the proposed confidence sampling, while fixing sampled tokens. A comparison between our proposed scheme and the aforementioned can be seen in Figure \ref{fig:masking}. 
An adjacent argument about the noising token can be made. Chang \etal use a unique mask token, about which we hypothesize is decreasing the diversity of outputs compared to using random tokens for the noising process. Denoising a specific masked image will always result in the exact same output logits. Diversity has to be forced during sampling, adding stochasticity through random sampling.
On the contrary, our proposed sampling inherently encourages diversity through the initial noise. In Figure~\ref{fig:masking} we show that a single denoising step, starting from full noise, results in very different outcomes. In a model trained with mask tokens, this prediction would always be the same.

\subsection{Token Predictor Architecture}
Our architecture consists of a U-Net-style \cite{https://doi.org/10.48550/arxiv.1505.04597}  encoder-decoder structure based on residual blocks \cite{resnet_he2016deep}, employing convolutional and attention in both, the encoder and the decoder path. The encoder and decoder consist of three levels. Attention blocks are only used at the two lowest levels to avoid memory overheads. To further increase the throughput, we use a patch size of 2 \cite{vit_dosovitskiy2020image,liu2022convnet}, which reduces the spatial dimensions while increasing the channels.
This allows the model to scale flexibly to arbitrary latent dimensions. % (Figure \ref{fig:arch}). 
Besides the latent image, every block takes in the conditional embeddings and the timestep embedding. We use cross-attention \cite{transformers_vaswani2017attention} to combine the conditional embeddings with the image latents. Multiple conditionings from different models, ByT5 and CLIP \cite{xue2022byt5,clip_radford2021learning}, are first projected into a shared latent space and afterwards concatenated and propagated into the cross-attention layers. To enable the model with more capabilities to learn from the CLIP embeddings, the pooled embedding is projected into four separate embeddings, making room for learning different aspects in the individual heads.

%Instead of using standard 2D convolutions, we are using depthwise convolutions, since these are significantly faster and consume much less memory \cite{https://doi.org/10.48550/arxiv.1704.04861,https://doi.org/10.48550/arxiv.1403.1687}. Furthermore, each block also contains a linear projection, mapping both conditional embeddings to the latent dimension. 

%This is followed by a channelwise convolution consisting of two fully connected layers combined via a GELU \cite{gelu_hendrycks2016gaussian} activation. Finally, we scale the activations by a learned constant and add the residual connection. Figure \ref{fig:arch} depicts the setup of our architecture visually. Note that our architecture is neither using attention, nor are we applying extensive normalization, making the model fast and memory-efficient.

\section{Experiments}
Prior to the training of the definitive model, we executed a series of scaled-down experiments to identify an appropriate and efficient architecture. This process involved training on reduced latent space dimensions, fewer model parameters, and smaller datasets.
Subsequent to these preliminary investigations, we proceeded to train our comprehensive model, called \textit{Paella}\footnote{We named the model after the popular food paella, as in our initial experiments generating food always was the first thing to work well.}. 
%We demonstrate our model's capability to perform text- and image-conditional image synthesis. %We empirically justify all of our design decisions via the ablation studies, demonstrating our model's capability to perform text-, and image-conditional image synthesis. We also report about other out-of-the-box applications of our model.

\subsection{Training}
\label{sec:paella_training}
%Determining the architecture for \textit{Paella} was conducted in a small-scale architecture search using the FFHQ (PLEASECITE) dataset. Additionally, 
We decided to use multiple models for conditioning and extract information from both text and images. The decision to use ByT5 \cite{xue2022byt5} was due to recent investigations between character-aware and character-blind \cite{liu2022character} models, which concluded the former to be better at text rendering tasks. Additionally including CLIP text- and image-embeddings opened up many interesting downstream applications such as style-transfer, mixing multiple images, guiding text-conditional generation via images, image variations etc.
Incorporating all previous findings and decisions, we trained our largest \textit{Paella} model with 1B parameters. It was trained on $900$ million images from the improved LAION-5B aesthetic \cite{schuhmann2022laion} dataset for $1M$ steps with a batch size of $2048$. We further finetuned the model on a subset of the aesthetic dataset with aesthetic scores higher than six for $200k$ steps using a batch size of $1024$. \textit{Paella} uses a ByT5-XL encoder and a \ac{CLIP} ViT-H/14 \cite{ilharco_gabriel_2021_5143773}. We trained on $128$ NVIDIA A100 @ 80GB for three weeks. All experiments use AdamW \cite{https://doi.org/10.48550/arxiv.1711.05101} for optimization with a learning rate of $1e^{-4}$ using a linear warm-up schedule for 30k steps.

\begin{figure}[t]
%\subfloat[][]{\includegraphics[width=0.5\textwidth]{figures/FID_sampling_steps.pdf}\label{fig:fid}}
%\subfloat[][]{\includegraphics[width=0.5\textwidth]{figures/CLIP_steps.pdf}\label{fig:clip_score}}
\centering\includegraphics[width=0.85\textwidth]{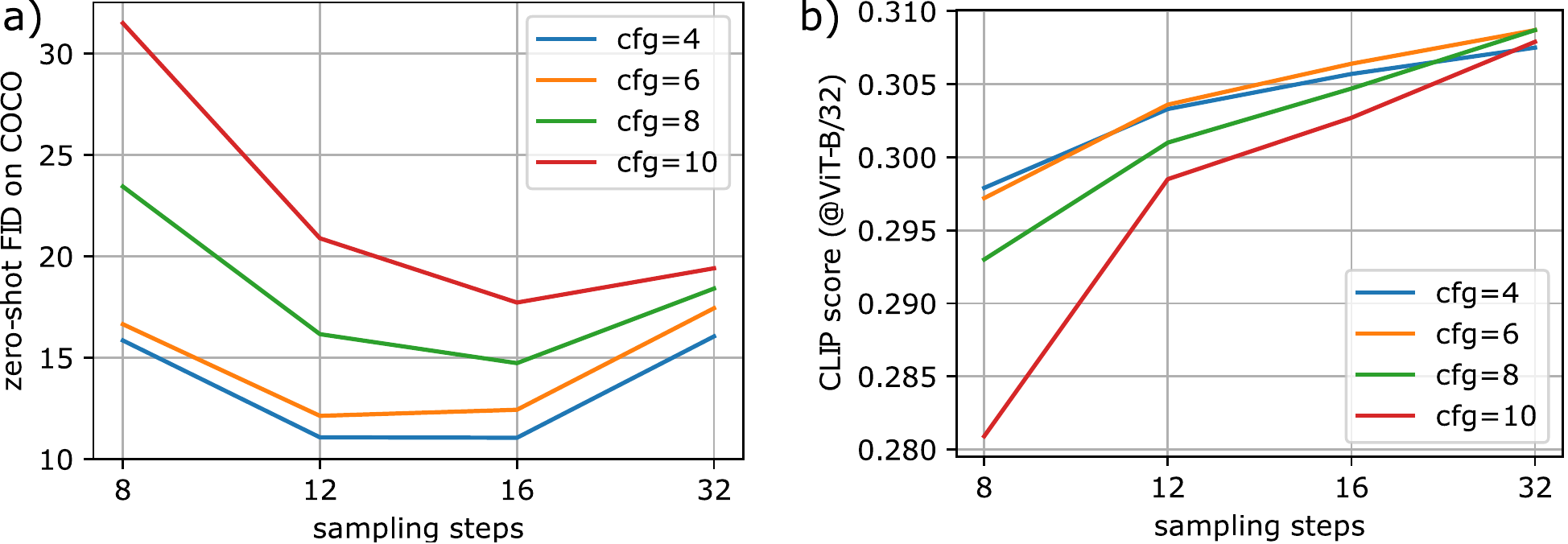}
\caption{Dependency of the a) zero-shot Fréchet Inception Distance (FID) score \cite{heusel2017gans} and the b) CLIP score on the number of sampling steps.}
\label{fig:fid_clip}
\end{figure}

\subsection{Text-Conditional Image Synthesis}
\label{sec:text_cond}
To demonstrate \textit{Paella's} text-conditional image generation capabilities we provide visual results (see Figure~\ref{fig:samples}), but also quantitative numerical analysis (see Table~\ref{tab:fid-comp}). We evaluated zero-shot \ac{FID} \cite{heusel2017gans} scores to determine the faithfulness and fidelity compared to ground truth images from MS COCO \cite{chen2015microsoft}. Next to \ac{FID} calculations, we assess the CLIP score between the captions and the generated images using CLIP ViT-B/32. We evaluate against different hyperparameter settings for the number of timesteps $T$ and \ac{CFG} weight $w$. The results can be found in Figure \ref{fig:fid_clip}. We find the \ac{FID} values to be highly competitive to state-of-the-art work as shown in Table \ref{tab:fid-comp}, while having a fraction of the parameters, compared to most models, and using an order of magnitude fewer sampling iterations. Moreover, the CLIP score results show an increase in alignment with more sampling steps and that smaller \ac{CFG} weights yield a better performance, especially for a lower number of iterations.

\begin{table}[]
\resizebox{\linewidth}{!}{
\begin{tabular}{|l|r|r|r|c|c|}
\hline
Model            & Parameters & Sampling Steps $\downarrow$ & FID-COCO-30k $\downarrow$  & open source & training data available   \\ \hline
CogView \cite{ramesh2021zero}           & 4B        & 1024            & 27.1 & \checkmark & (partial) \\ 
Parti \cite{https://doi.org/10.48550/arxiv.2206.10789}          & 20B        & 1024            & \textbf{7.23} & -- & -- \\ 
Make-A-Scene \cite{gafni2022make}     & 4B         & 1024           & 11.84       & -- & (partial) \\ 
Imagen \cite{saharia2022photorealistic}           & 2B         & 1000          & 7.27   & -- & --\\ 
DALL-E \cite{ramesh2021zero}           & 12B        & 256            & 17.89 & -- & --\\
DALL-E 2 \cite{dalle2_ramesh2022hierarchical}         & 3.5B       & 250            & 10.39  & -- & --\\ 
GLIDE \cite{glide_nichol2021glide}            & 3.5B       & 250            & 12.24   & -- & -- \\ 

LDM \cite{rombach2022high} & 0.4B & 250 & 12.63 & \checkmark & \checkmark \\
MUSE-3B \cite{chang2023muse} & 3B & 24 & 7.78 & -- & -- \\
%Stable-Diffusion v1.4 @ $256\times 256$ \cite{rombach2022high} & 860M       & 50             & 25.40* & \checkmark\\ 
%\hline
%Paella wo/CFG & \multirow{2}{*}{\textbf{1B}} & \multirow{2}{*}{\textbf{12}} & ? & \multirow{2}{*}{\checkmark} \\
Paella  (proposed) &    1B     &     \textbf{12}       & 11.07   &  \checkmark & \checkmark \\ 
\hline
\end{tabular}
}
\caption{Comparison of the zero-shot Fréchet Inception Distance to other state-of-the-art text-to-image methods on $256\times256$ images.} % , including an ablation study of our \textit{Paella} model with (w/) and without (wo/) classifier-free guidance (CFG). $^*$: own evaluation, generated with 50 DDIM \cite{song2020denoising} steps and $w=7.5$ at $512\times 512$, downsampled to $256\times 256.$
\label{tab:fid-comp}
\end{table}

%However, our model has the great benefit of needing a fraction of sampling steps in contrast to other methods. In combination with the inference speed of each step (due to the use of the CNN), \textit{Paella} can sample high quality images in before unseen times. In Figure \ref{fig:inferencetime} we show inference speeds given different batch sizes on a single NVIDIA A100. The model is able to sample 48 images in 2.94 seconds.
%After finding suitable values for $T$ and $w$ we compare if \ac{CFG} is beneficial or detrimental for image generations. The results of this ablation study can be seen in Table \ref{tab:fid-comp} as well.

\section{Discussion \& Limitations}
This work introduces an enhanced scheme for sampling in quantized latent spaces for text-to-image applications, allowing for a reduced number of sampling steps, while showing competitive fidelity, as measured by the \ac{FID}, to state-of-the-art works. 
% We found that our model Paella, which we also introduced in this work, does not beat the most recent approaches by the major industrial research labs, such as Parti, Imagen or MUSE in terms of the \ac{FID}. However, compared to other openly accessible approaches using openly available datasets (e.g., LDM \cite{rombach2022high}), our approach requires less sampling steps while achieving a more favorable \ac{FID} score (see Table \ref{tab:fid-comp}). 
As shown in Figure \ref{fig:fid_clip}, the \ac{FID} value of our model reaches its minimum for only 12 sampling steps, which in our view is a direct consequence of the novel sampling scheme, allowing for a higher diversity in the images, compared to the masking strategies employed by approaches such as MUSE or MaskGIT.

One interesting finding of our research is that our model is not particularly good at rendering text in images, which was reported by \cite{liu2022character} and was the main motivation for choosing ByT5 over T5. We link this possibly to an unintended consequence resulting from using quantized tokens, which only allows to either fully destroy or preserve information of a token. We leave this research question open for future investigation.

Another intriguing observation is that the model performance does not correlate positively with an increase in sampling steps regarding FID evaluations. Instead, it shows optimal performance at approximately 12 inference steps. This observation corroborates the findings of MaskGIT \cite{maskgit_chang2022maskgit} and contradicts the conventional heuristic in diffusion models, where a higher number of sampling steps is typically associated with improved performance.
However regarding the CLIP score, the model performs consistently better using more inference steps. This observation might indicate that fidelity emerges earlier than conditional alignment, showing that it is easier for the model to generate visually appealing images, than making them well aligned with the prompts.
Contradicting to previous work as well, a higher \ac{CFG} weight does not lead to an increase in the CLIP score for \textit{Paella} and rather smaller guidance weights outperform higher ones on all timesteps.
%ByT5 did not work as good as expected. Generating text in images did not work reliably. We think this is caused by ...

%One central limitation of our evaluation is the parameter and training steps difference between \textit{Paella} and other state-of-the-art models. The amount of images other models have seen during training outweighs our experiments by magnitudes, which makes fair comparisons hard, especially when many of these models are kept private. To this degree, we hope to make a contribution to reproducible and transparent science in our field by providing the complete model including all weights.

%{\color{red}
%- stable diffusion comparison is hard
%- statement for ethical usage / broader impact -> say that humanity needs other ways than just centralizing
%- We need some discussion on the high FID / low precision \& recall, e.g.:
%--> State that we found FID values comparable to SOTA methods, but only when evaluated on LAION-30k.
%--> We assume that this stems from a considerable distribution shift (why?)
%--> Potentially could also reside distribution used for training the VQGAN
%--> Diffusion-based models perform continuous noising at pixel level, while we do discrete noising, each noised token looses 100% of information. This limits the applications, like smooth transitions between domains.
%--> Our method does not work on a pixel level, but using a low compression. --> Future work
%--> Gamma-Function: We just picked the one that was experimentally working on MaskGIT --> Future Work
%}

\section{Conclusion}
In this work we presented \textit{Paella}, a text- and image-conditional image generation system using a novel training objective and an improved sampling strategy. We showed that our model can generate high-fidelity images despite being smaller and requiring less steps for sampling than existing models, while still reaching competitive numerical results. We especially want to highlight the simple and straightforward setup of this model regarding training and sampling compared to models based on diffusion or transformers and believe this method will make generative techniques more accessible to a variety of people, even outside the research field of generative AI, which we argue will become crucial as this technology progresses further. 
We provide the final model weights and code on github \footnote{\url{https://github.com/delicious-tasty/Paella}}. Further, we provide training scripts and inference notebooks to support reproducibility of our findings.

%\section*{Acknowledgements}

%Acknowledgements withheld for review.
%The authors wish to express their thanks to Stability AI Inc. for providing generous computational resources for our experiments and LAION gemeinnütziger e.V. for dataset access and support.

\begin{acronym}
\acro{CLIP}[CLIP]{Contrastive Language-Image Pretraining}
\acro{MaskGIT}[MaskGIT]{Masked Generative Image Transformer}
\acro{GAN}[GAN]{Generative Adversarial Network}
\acro{VQ-VAE}[VQ-VAE]{Vector-quantized Variational Autoencoder}
\acro{VQGAN}[VQGAN]{Vector-quantized Generative Adversarial Network}
\acro{CFG}[CFG]{Classifier-Free Guidance}
%\acro{LTS}[LTS]{Locally Typical Sampling}
\acro{FID}[FID]{Fréchet Inception Distance}
\end{acronym}

{\small
\bibliographystyle{apalike}
\bibliography{egbib}

\begin{thebibliography}{}

\bibitem[Chang et~al., 2023]{chang2023muse}
Chang, H. et~al. (2023).
\newblock Muse: Text-to-image generation via masked generative transformers.
\newblock {\em arXiv:2301.00704}.

\bibitem[Chang et~al., 2022]{maskgit_chang2022maskgit}
Chang, H., Zhang, H., Jiang, L., Liu, C., and Freeman, W.~T. (2022).
\newblock {MaskGIT}: Masked generative image transformer.
\newblock In {\em Proceedings of the IEEE/CVF Conference on Computer Vision and
  Pattern Recognition}, pages 11315--11325.

\bibitem[Chen et~al., 2015]{chen2015microsoft}
Chen, X., Fang, H., Lin, T.-Y., Vedantam, R., Gupta, S., Doll{\'a}r, P., and
  Zitnick, C.~L. (2015).
\newblock Microsoft {COCO} captions: Data collection and evaluation server.
\newblock {\em arXiv:1504.00325}.

\bibitem[Dhariwal and Nichol, 2021]{dhariwal2021diffusion}
Dhariwal, P. and Nichol, A. (2021).
\newblock Diffusion models beat gans on image synthesis.
\newblock {\em Advances in Neural Information Processing Systems},
  34:8780--8794.

\bibitem[Ding et~al., 2021]{ding2021cogview}
Ding, M., Yang, Z., Hong, W., Zheng, W., Zhou, C., Yin, D., Lin, J., Zou, X.,
  Shao, Z., Yang, H., et~al. (2021).
\newblock Cogview: Mastering text-to-image generation via transformers.
\newblock {\em Advances in Neural Information Processing Systems},
  34:19822--19835.

\bibitem[Ding et~al., 2022]{ding2022cogview2}
Ding, M., Zheng, W., Hong, W., and Tang, J. (2022).
\newblock Cogview2: Faster and better text-to-image generation via hierarchical
  transformers.
\newblock {\em arXiv:2204.14217}.

\bibitem[Dosovitskiy et~al., 2020]{vit_dosovitskiy2020image}
Dosovitskiy, A., Beyer, L., Kolesnikov, A., Weissenborn, D., Zhai, X.,
  Unterthiner, T., Dehghani, M., Minderer, M., Heigold, G., Gelly, S., et~al.
  (2020).
\newblock An image is worth 16x16 words: Transformers for image recognition at
  scale.
\newblock In {\em Proceedings of the International Conference on Learning
  Representations (ICLR)}.

\bibitem[Esser et~al., 2021]{vqgan_esser2021taming}
Esser, P., Rombach, R., and Ommer, B. (2021).
\newblock Taming transformers for high-resolution image synthesis.
\newblock In {\em Proceedings of the IEEE/CVF Conference on Computer Vision and
  Pattern Recognition}, pages 12873--12883.

\bibitem[Gafni et~al., 2022]{gafni2022make}
Gafni, O., Polyak, A., Ashual, O., Sheynin, S., Parikh, D., and Taigman, Y.
  (2022).
\newblock Make-a-scene: Scene-based text-to-image generation with human priors.
\newblock {\em arXiv:2203.13131}.

\bibitem[He et~al., 2016]{resnet_he2016deep}
He, K., Zhang, X., Ren, S., and Sun, J. (2016).
\newblock Deep residual learning for image recognition.
\newblock In {\em Proceedings of the IEEE Conference on Computer Vision and
  Pattern Recognition}, pages 770--778.

\bibitem[Heusel et~al., 2017]{heusel2017gans}
Heusel, M., Ramsauer, H., Unterthiner, T., Nessler, B., and Hochreiter, S.
  (2017).
\newblock {GANs} trained by a two time-scale update rule converge to a local
  nash equilibrium.
\newblock {\em Advances in Neural Information Processing Systems}, 30.

\bibitem[Ho et~al., 2020]{ho2020denoising}
Ho, J., Jain, A., and Abbeel, P. (2020).
\newblock Denoising diffusion probabilistic models.
\newblock {\em Advances in Neural Information Processing Systems},
  33:6840--6851.

\bibitem[Ho and Salimans, 2022]{classifier_free_guidance_ho2022classifier}
Ho, J. and Salimans, T. (2022).
\newblock Classifier-free diffusion guidance.
\newblock {\em arXiv:2207.12598}.

\bibitem[Ilharco et~al., 2021]{ilharco_gabriel_2021_5143773}
Ilharco, G., Wortsman, M., Carlini, N., Taori, R., Dave, A., Shankar, V.,
  Namkoong, H., Miller, J., Hajishirzi, H., Farhadi, A., and Schmidt, L.
  (2021).
\newblock {\em OpenCLIP}.
\newblock Zenodo.

\bibitem[Karras et~al., 2022]{https://doi.org/10.48550/arxiv.2206.00364}
Karras, T., Aittala, M., Aila, T., and Laine, S. (2022).
\newblock Elucidating the design space of diffusion-based generative models.

\bibitem[Liu et~al., 2022a]{https://doi.org/10.48550/arxiv.2202.09778}
Liu, L., Ren, Y., Lin, Z., and Zhao, Z. (2022a).
\newblock Pseudo numerical methods for diffusion models on manifolds.
\newblock {\em arXiv:2202.09778}.

\bibitem[Liu et~al., 2022b]{liu2022character}
Liu, R., Garrette, D., Saharia, C., Chan, W., Roberts, A., Narang, S., Blok,
  I., Mical, R., Norouzi, M., and Constant, N. (2022b).
\newblock Character-aware models improve visual text rendering.
\newblock {\em arXiv:2212.10562}.

\bibitem[Liu et~al., 2022c]{liu2022convnet}
Liu, Z., Mao, H., Wu, C.-Y., Feichtenhofer, C., Darrell, T., and Xie, S.
  (2022c).
\newblock A convnet for the 2020s.
\newblock In {\em Proceedings of the IEEE/CVF Conference on Computer Vision and
  Pattern Recognition}, pages 11976--11986.

\bibitem[Loshchilov and Hutter,
  2019]{https://doi.org/10.48550/arxiv.1711.05101}
Loshchilov, I. and Hutter, F. (2019).
\newblock Decoupled weight decay regularization.
\newblock {\em International Conference on Learning Representations (ICLR)}.

\bibitem[Nichol et~al., 2021]{glide_nichol2021glide}
Nichol, A., Dhariwal, P., Ramesh, A., Shyam, P., Mishkin, P., McGrew, B.,
  Sutskever, I., and Chen, M. (2021).
\newblock Glide: Towards photorealistic image generation and editing with
  text-guided diffusion models.
\newblock {\em arXiv:2112.10741}.

\bibitem[Radford et~al., 2021]{clip_radford2021learning}
Radford, A., Kim, J.~W., Hallacy, C., Ramesh, A., Goh, G., Agarwal, S., Sastry,
  G., Askell, A., Mishkin, P., Clark, J., et~al. (2021).
\newblock Learning transferable visual models from natural language
  supervision.
\newblock In {\em International Conference on Machine Learning}, pages
  8748--8763. PMLR.

\bibitem[Raffel et~al., 2020]{raffel2020exploring}
Raffel, C., Shazeer, N., Roberts, A., Lee, K., Narang, S., Matena, M., Zhou,
  Y., Li, W., Liu, P.~J., et~al. (2020).
\newblock Exploring the limits of transfer learning with a unified text-to-text
  transformer.
\newblock {\em J. Mach. Learn. Res.}, 21(140):1--67.

\bibitem[Ramesh et~al., 2022]{dalle2_ramesh2022hierarchical}
Ramesh, A., Dhariwal, P., Nichol, A., Chu, C., and Chen, M. (2022).
\newblock Hierarchical text-conditional image generation with {CLIP} latents.
\newblock {\em arXiv:2204.06125}.

\bibitem[Ramesh et~al., 2021]{ramesh2021zero}
Ramesh, A., Pavlov, M., Goh, G., Gray, S., Voss, C., Radford, A., Chen, M., and
  Sutskever, I. (2021).
\newblock Zero-shot text-to-image generation.
\newblock In {\em International Conference on Machine Learning}, pages
  8821--8831. PMLR.

\bibitem[Reed et~al., 2016]{reed2016generative}
Reed, S., Akata, Z., Yan, X., Logeswaran, L., Schiele, B., and Lee, H. (2016).
\newblock Generative adversarial text to image synthesis.
\newblock In {\em International conference on machine learning}, pages
  1060--1069. PMLR.

\bibitem[Rombach et~al., 2022]{rombach2022high}
Rombach, R., Blattmann, A., Lorenz, D., Esser, P., and Ommer, B. (2022).
\newblock High-resolution image synthesis with latent diffusion models.
\newblock In {\em Proceedings of the IEEE/CVF Conference on Computer Vision and
  Pattern Recognition}, pages 10684--10695.

\bibitem[Ronneberger et~al., 2015]{https://doi.org/10.48550/arxiv.1505.04597}
Ronneberger, O., Fischer, P., and Brox, T. (2015).
\newblock U-net: Convolutional networks for biomedical image segmentation.
\newblock In {\em International Conference on Medical Image Computing and
  Computer-Assisted Intervention}, pages 234--241. Springer.

\bibitem[Saharia et~al., 2022]{saharia2022photorealistic}
Saharia, C., Chan, W., Saxena, S., Li, L., Whang, J., Denton, E., Ghasemipour,
  S. K.~S., Ayan, B.~K., Mahdavi, S.~S., Lopes, R.~G., et~al. (2022).
\newblock Photorealistic text-to-image diffusion models with deep language
  understanding.
\newblock {\em arXiv:2205.11487}.

\bibitem[Schuhmann et~al., 2022]{schuhmann2022laion}
Schuhmann, C., Beaumont, R., Vencu, R., Gordon, C., Wightman, R., Cherti, M.,
  Coombes, T., Katta, A., Mullis, C., Wortsman, M., et~al. (2022).
\newblock Laion-5b: An open large-scale dataset for training next generation
  image-text models.
\newblock {\em arXiv:2210.08402}.

\bibitem[Sohl-Dickstein et~al.,
  2015]{https://doi.org/10.48550/arxiv.1503.03585}
Sohl-Dickstein, J., Weiss, E., Maheswaranathan, N., and Ganguli, S. (2015).
\newblock Deep unsupervised learning using nonequilibrium thermodynamics.
\newblock In {\em International Conference on Machine Learning}, pages
  2256--2265. PMLR.

\bibitem[Vaswani et~al., 2017]{transformers_vaswani2017attention}
Vaswani, A., Shazeer, N., Parmar, N., Uszkoreit, J., Jones, L., Gomez, A.~N.,
  Kaiser, {\L}., and Polosukhin, I. (2017).
\newblock Attention is all you need.
\newblock {\em Advances in Neural Information Processing Systems}, 30.

\bibitem[Xue et~al., 2022]{xue2022byt5}
Xue, L., Barua, A., Constant, N., Al-Rfou, R., Narang, S., Kale, M., Roberts,
  A., and Raffel, C. (2022).
\newblock Byt5: Towards a token-free future with pre-trained byte-to-byte
  models.
\newblock {\em Transactions of the Association for Computational Linguistics},
  10:291--306.

\bibitem[Yu et~al., 2022]{https://doi.org/10.48550/arxiv.2206.10789}
Yu, J., Xu, Y., Koh, J.~Y., Luong, T., Baid, G., Wang, Z., Vasudevan, V., Ku,
  A., Yang, Y., Ayan, B.~K., Hutchinson, B., Han, W., Parekh, Z., Li, X.,
  Zhang, H., Baldridge, J., and Wu, Y. (2022).
\newblock Scaling autoregressive models for content-rich text-to-image
  generation.
\newblock {\em arXiv:2206.10789}.

\bibitem[Zhang et~al., 2017]{zhang2017stackgan}
Zhang, H., Xu, T., Li, H., Zhang, S., Wang, X., Huang, X., and Metaxas, D.~N.
  (2017).
\newblock Stackgan: Text to photo-realistic image synthesis with stacked
  generative adversarial networks.
\newblock In {\em Proceedings of the IEEE international conference on computer
  vision}, pages 5907--5915.

\end{thebibliography}
}

\end{document}